\documentclass[runningheads]{llncs}
\usepackage{graphicx}
\usepackage[utf8]{inputenc} 
\usepackage[T1]{fontenc}    
\usepackage{hyperref}       
\usepackage{url}            
\usepackage{booktabs}       
\usepackage{amsfonts}       
\usepackage{nicefrac}       
\usepackage{microtype}      
\usepackage{lipsum}
\usepackage{graphicx}
\usepackage{cite}
\graphicspath{ {./images/} }
\usepackage{amsmath}
\usepackage{amssymb}
\usepackage{todonotes}
\usepackage{breqn}
\usepackage{colortbl}
\usepackage{multirow}
\usepackage{arydshln}
\usetikzlibrary{arrows}
\usetikzlibrary{shapes}
\usepackage{authblk}
\usepackage{bm}
\usepackage{versions}
\usepackage[bbgreekl]{mathbbol}
\usepackage{textgreek}
\usepackage{apxproof}
\usepackage{multicol}
\newenvironment{ijcai}{\color{black}}{}
\newenvironment{arxiv}{\color{black}}{}

\excludeversion{ijcai}

\def\mc{\mbox{\sc mc}}

\def\L{\mathcal{L}}
\def\bx{\bm{x}}

\def\fomc{\mbox{\sc fomc}}

\def\wfomc{\mbox{\sc wfomc}}
\def\preds{\mathcal{P}}
\def\R{\mathbb{R}}

\def\bh{\bm{h}}
\def\kh{\bk,\bh}

\def\CC{\mathbb{C}}
\def\bk{{\bm{k}}}

\def\cgreen{\cellcolor{green!10}}
\def\cred{\cellcolor{red!10}}
\def\card#1{{\left|#1\right|}}

\newif\ifhidesm

\ifhidesm
\usepackage{environ}
\NewEnviron{hide}{}

\fi

\begin{document}

\title{Weighted Model Counting in FO$^2$ with Cardinality Constraints : A Closed Form Formula}
%
\titlerunning{WFOMC: A closed form formula}
%
\author{Sagar Malhotra\inst{1,2} \and
Luciano Serafini\inst{1}}
\authorrunning{S. Malhotra et al.}
%
\institute{University of Trento,Italy\\
 \and
 Fondazione Bruno Kessler, Italy\\
 \email{\{smalhotra,serafini@\}@fbk.eu}}
\maketitle              
\begin{abstract}
  Weighted First-Order Model Counting (WFOMC) computes the weighted sum of the models of a 
  first-order theory on a  given finite domain. 
  WFOMC has emerged as a fundamental tool for probabilistic inference.
\begin{arxiv}
  Algorithms for WFOMC that run in polynomial time w.r.t. the domain size are called lifted inference algorithms. 
  Such algorithms have been developed for multiple extensions of
  FO$^2$ (the fragment of first-order logic with two variables) for
  the special case of symmetric weight functions. 
\end{arxiv}
  We introduce the concept of \emph{lifted interpretations} as a tool for formulating polynomials for WFOMC.
  Using lifted interpretations, we reconstruct the closed-form formula for polynomial-time FOMC in the universal fragment of FO$^2$, earlier proposed by Beame et al.  
  We then expand this closed-form to incorporate existential quantifiers and  cardinality constraints without losing domain-liftability. 
  Finally, we show that the obtained closed-form motivates a natural definition of  a family of weight functions strictly larger 
  than symmetric weight functions.  

\end{abstract}


\section{Introduction}
%
%
\begin{arxiv}
Statistical Relational Learning (SRL) attempts to reason about probabilistic
distributions over properties of relational domains \cite{SRL_LISA,SRL_LUC}.
Most SRL frameworks use formulas in a logical language to provide a compact
representation of the domain structure. Probabilistic knowledge on relational
domain can be specified by assigning a weight to every interpretation of the
logical language. One of the advantages of this approach
is that probabilistic inference can be cast as Weighted Model
Counting \cite{CHAVIRA2008772}.
\end{arxiv}
First-Order Logic (FOL) allows specifying structural knowledge with formulas that
contain individual variables that range over all the individuals of the domain.
Probabilistic inference on domains described in FOL requires the grounding (aka
instantiation) of all the individual variables with all the occurrences of the
domain elements. This grounding leads to an exponential blow up of the
complexity of the model description and hence the probabilistic inference. 

\emph{Lifted inference}
\cite{First_Order_Prob_Inf,de_Salvo} aims at resolving
this problem by exploiting symmetries inherent to the FOL structures. In recent years, \emph{Weighted First-Order Model Counting}
has emerged as a useful formulation for probabilistic inference in statistical relational learning frameworks \begin{ijcai}\cite{SRL_LISA,SRL_LUC}\end{ijcai}.
Formally, WFOMC \begin{ijcai}\cite{CHAVIRA2008772}\end{ijcai} refers to the task of calculating the weighted sum of the models of
a formula $\Phi$ over a domain of a finite size
\begin{ijcai}
$\wfomc(\Phi,w,n) = \sum_{\omega\models\Phi}w(\omega)$,
\end{ijcai}
\begin{arxiv}
$$\wfomc(\Phi,w,n) = \sum_{\omega\models\Phi}w(\omega)$$
\end{arxiv}
where $n$ is the 
cardinality of the domain and $w$ is a \emph{weight function}
that associates a real number to each interpretation $\omega$. 
FOL theories $\Phi$ and weight functions $w$ which admit an algorithm that
computes $\wfomc(\Phi,w,n)$ in a polynomial time w.r.t. $n$ are called
\emph{domain-liftable} \cite{LIFTED_PI_KB_COMPLETION}.

In the past decade, multiple extensions of FO$^2$ (the fragment of
FOL with two variables) have been proven to be
domain-liftable
\cite{gogate2016probabilistic,DOMAIN_RECURSION,kazemi2016new,kuusisto2018weighted,kuzelka2020weighted}. 
\begin{arxiv}
These results are formulated over a special class of weight functions known as \emph{symmetric weight functions} \cite{Symmetric_Weighted} and utilise lifted inference rules which are able to exploit the symmetry of FOL formulas in a rule based manner.
\end{arxiv}

In this paper instead of relying on an algorithmic approach to WFOMC,
as in \cite{LIFTED_PI_KB_COMPLETION}, our objective is to find a
closed-form for WFOMC in FO$^2$ that can be easily extended to larger
classes of first-order formulas. To this aim we introduce the novel
notion of \emph{lifted interpretation}: a completely first-order
concept 
\begin{arxiv}independent of the domain\end{arxiv}.
Lifted interpretations allows us to reconstruct the closed-form
formula for First Order Model Counting (FOMC) in FO$^2$ proposed in \cite{Symmetric_Weighted} and
to extend it to larger classes of FO formulas.  
We see the following key benefits of the presented formulation:
\begin{enumerate}
\item \emph{The formula is easily extendeds to FO$^2$ with cardinality
    constraints without losing domain-liftability}. A cardinality constraint on
  an interpretation is a constraint on the number of elements for
  which a certain predicate holds. Earlier approaches to dealing with 
  cardinality constraints involves either using Discrete Fourier Transform \cite{kuzelka2020lifted_DFT} over 
  complex numbers or evaluating lagrange interpolation \cite{kuzelka2020weighted}. 

\item \emph{The formula deals with equality in constant time w.r.t the domain cardinality}. Previous works in 
WFOMC \cite{Symmetric_Weighted} require additional $n+1$ calls to the WFOMC oracle, where $n$ is the domain cardinality. 
\item \emph{The proposed formula solves the model counting problem
    without introducing (negative) weights}.  This has the advantage
  of allowing separate treatment for model counting from weighted model
  counting. Furthermore, though it has been shown that WFOMC is well
  defined also with negative weights, the connections between negative
  weights and probability turns out to be less clean, as it leads to
  negative and larger than one probability values \cite{kazemi2017domain}.
\item \emph{The formula computes WFOMC for a class of weight functions
    strictly larger than symmetric weight functions.}  This extended
  class of weight functions allow to model the recently introduced
  count distributions \cite{complexMLNkuzelka2020}.
\end{enumerate}

Most of the paper focuses on FOMC. We
then show how weighted model counting can be obtained by multiplying
each term of the resulting formula for FOMC with the corresponding
weight. This allows us to separate the treatment of the counting part
from the weighting part. The paper is therefore structured as follows:
The next section describes the related work in the literature on
WFOMC. We then present our formulation of closed-form formula for FOMC
given in \cite{Symmetric_Weighted} for the universally quantified fragment
of FO$^2$. We then extend this formula to incorporate cardinality constraints. 
In the successive section, we show how this formula can
be used to compute FOMC also in the presence of existential
quantifiers. The last part of the paper extends
the formula for FOMC to WFOMC for the case of symmetric weight
functions and for a larger class of weight functions that allow to model 
count distributions \cite{complexMLNkuzelka2020}.


\section{Related work}
Weighted First Order Model Counting (WFOMC) was initially defined in \cite{LIFTED_PI_KB_COMPLETION}. 
The paper provides an algorithm for WFOMC over universally quantified theories based on a \emph{knowledge compilation} technique, 
which transforms an FOL theory to a  \emph{first order deterministic decomposable
normal form (FO d-DNNF)}\footnote{FOL-d-DNNF is a d-DNNF \cite{darwiche2002knowledge} where literals may contain individual variables}. 
  A successive paper \cite{DOMAIN_RECURSION} has formalized the notion of \emph{domain lifted theory} 
  i.e. a first order theory for which WFOMC can be computed in polynomial time in the size of the domain. 
  The same paper shows that a theory composed of a set of universally quantified clauses containing at most two variables is domain liftable. 
  This is done by combining knowledge compilation with a technique called \emph{domain recursion}. 
  In domain recursion, WFOMC of a theory on a domain of $n$ elements is rewritten in terms of WFOMC of the same theory in a domain with $n-1$ elements, 
  by partially grounding the theory with a single element of the domain. 
  A successive paper \cite{broeck2013} extends this procedure to theories in full $\mathrm{FO^{2}}$ (i.e, where existential quantification is allowed) 
  by applying skolemization to remove existentially quantified variables.  
  The major drawback of these technique is that it introduces negative weights, 
  and therefore it makes it more complex to use it for probabilistic inference which requires non-negative weights. 
  These results are theoretically analysed in \cite{Symmetric_Weighted}, which provides a closed-form formula for WFOMC in FO$^2$. 
  \cite{kuusisto2018weighted} extends the domain liftability results to FO$^2$ with a functionality axiom, 
  and for sentences in \emph{uniform one-dimensional fragment} U$_1$ \cite{kieronski2015uniform}. 
  It also proposes a closed-form formula for WFOMC in $\mathrm{FO^{2}}$ with functionality constraints. 
  \cite{kuzelka2020weighted} recently proposed a uniform treatment of WFOMC for FO$^2$ with cardinality constraints  and counting quantifiers, 
  proving these theories to be domain-liftable.  
  Finally, \cite{kazemi2017domain} re-investigates the problem of skolemization arguing that negative weights can be prohibitive and that the skolemization procedure is computationally expensive. 
  The paper gives examples of theories for which skolemization can be bypassed using domain recursion. 
  With respect to the state of the art approaches to WFOMC, 
  we propose an approach that provides a closed-form for WFOMC with cardinality constraints from which the PTIME complexity is immediately evident. 
  Importantly, this doesn't require the introduction of negative weights. 
  Furthermore, w.r.t. the closed-form proposed in \cite{kuusisto2018weighted} and \cite{Symmetric_Weighted}, 
  our proposal for FOMC does not use weights, keeping the counting and the weighting part separate. 
  Finally, \cite{complexMLNkuzelka2020} introduces Complex Markov Logic Networks, 
  which use complex-valued weights and allow for full expressivity over a class of distributions called \emph{count distributions}. 
  We show in the last section of the paper that our formalization is complete w.r.t. this class of distributions.



\section{FOMC for Universal Formulas}
Let $\L$ be a first-order function free language with equality. 
A \emph{pure universal formula} in $\mathcal{L}$ is a formula of the form
\begin{align}
  \label{eq:universal-formula}
\forall x_1 \dots \forall x_m.\Phi(x_1,\dots,x_m)
\end{align}
where $X=\{x_1,\dots,x_m\}$ is a set of $m$ distinct variables occurring
in $\Phi(x_1,\dots,x_m)$, and $\Phi(x_1,\dots,x_m)$ is a quantifier
free formula that does not contain any constant symbol. We use the
compact notation $\Phi(\bx)$ for $\Phi(x_1,\dots,x_m)$, where
$\bx=(x_1,\dots,x_{m})$. Notice that we distinguish between the
$m$-tuple of variables $\bx$ and the \emph{set} of variables denoted
by $X$. For every $\bm{\sigma}=(\sigma_1,\dots,\sigma_m)$,
$m$-tuple of constants or variables, 
$\Phi(\bm\sigma)$ denotes the
result of uniform substitution of $x_i$ with $\sigma_i$ in
$\Phi(\bx)$. 
If $\Sigma\subseteq X\cup C$ is the set of constants or variables of
$\L$ and $\Phi(\bx)$ a pure universal formula then $\Phi(\Sigma)$
denotes the formula:
\begin{equation}
\label{Phi_set}
\Phi(\Sigma) = \bigwedge_{\bm\sigma\in\Sigma^{m}}\Phi(\bm\sigma)
\end{equation}

\begin{lemma}
For any arbitrary pure universal formula  $\forall \bx \Phi(\bx)$, the following equivalence holds: 
\begin{equation}
    \forall \bx \Phi(\bx) \leftrightarrow \forall \bx  \Phi(X)
\end{equation}
\end{lemma}

\begin{proof}
  For any $ \bm{x'}\in X^{m}$, we have that 
  $\forall \bm{x} \Phi(\bm{x}) \rightarrow \forall
  \bm{x}\Phi(\bm{x'})$ is valid. Which implies that 
  $\forall \bm{x} \Phi(\bm{x}) \rightarrow \bigwedge_{\bx'\in X^m}\forall
  \bm{x}\Phi(\bm{x'})$ is also valid. Since $\forall$ and $\wedge$ commute,
  we have that $\forall \bx.\Phi(\bx) \rightarrow
  \forall\bx.\Phi(X)$. The viceversa is obvious since $\Phi(\bx)$ is
  one of the conjuncts in $\Phi(X)$. 
\end{proof}

\begin{example}
  \label{ex:running}
  Let $\Phi(x,y) = A(x) \land R(x,y) \land x\neq y\rightarrow A(y)$,
  then $\Phi(X=\{x,y\})$ is the following formula 
\begin{align}
\begin{array}{l@{\ }l} 
(A(x) \land R(x,x)\land x\neq x\rightarrow A(x)) &\land (A(x) \land R(x,y)\land x\neq y  \rightarrow A(y))\land \\
(A(y) \land R(y,x)\land y\neq x\rightarrow A(x))  &\land  (A(y) \land R(y,y)\land y\neq y \rightarrow A(y))
\end{array}
\label{eq:example}
\end{align}
\end{example}
Notice that in $\Phi(X)$ we can assume that two distinct variables $x$
and $y$ are grounded to different domain elements. Indeed, the cases
in which $x$ and $y$ are grounded to the same domain element is taken
into account by the conjunct in which $y$ is replaced by $x$.  See for
instance the first and the last conjunct of \eqref{eq:example}.

\begin{definition}[Lifted interpretation]
A \emph{lifted interpretation} $\tau$ of a quantifier free
formula $\Phi(\bx)$ is a function that assigns to each atom of
$\Phi(X)$ either $0$ or $1$ ($0$ means false and $1$ true) and
assigns 1 to $x_i=x_i$ and 0 to $x_i=x_j$ if $i\neq j$. 
\end{definition}

Lifted interpretations allow associating truth values to
pure universal formulas. The 
truth value of $\Phi(\bx)$ under the truth assignment $\tau$, denoted
by $\tau(\Phi(\bx))$, is obtained by applying the classical
propositional logic of the connectives.  Notice that $\tau$ is not an 
FOL interpretation as it assigns truth values to atoms that contain 
free variables, and not to their groundings. 

\begin{example}
  \label{ex:truth-assignment}
Following is the  example of a lifted interpretation for the formula \eqref{eq:example} of Example~\ref{ex:running}:
$$
\begin{array}{lcccccc} 
 & A(x) & R(x,x) & A(y) & R(y,y) & R(x,y) & R(y,x) \\ \hline
  \tau & \cgreen 0 & \cgreen1 & \cred1 & \cred1 & \cgreen0 & \cred1 \\
  & \multicolumn{2}{c}{\tau_x} 
       & \multicolumn{2}{c}{\tau_y}
   & \multicolumn{2}{c}{\tau_{xy}} 
 \end{array}
$$
We omit the truth assignments of equality atoms, since it is fixed. We
have that $\tau(\eqref{eq:example}) = 0$.
\end{example}
As highlighted in the previous example, 
any lifted interpretation $\tau$ can be split into a set of partial
lifted interpretations $\tau_{X'}$, where $X'\subseteq X$ is a 
non-empty subset of variables occurring in $\Phi$. In the example
$X=\{x,y\}$ and $\tau_{\{x\}}$ (simply denoted
 by $\tau_x$) contains the assignments to the atoms containing only
  $x$ and we can similarly define $\tau_y$. We also have $\tau_{\{x,y\}}$, written as
  $\tau_{xy}$, containing the assignments to the atoms that contain both
  $x$ and $y$. 



\begin{arxiv}
\begin{example}
Consider the  assignment of example \ref{ex:truth-assignment} and the
one obtained by the permutation $\pi$ that exchanges $x$ and $y$

$$
\begin{array}{lcccccc} 
& A(x) & R(x,x) & A(y) & R(y,y) & R(x,y) & R(y,x) \\ \hline
  \tau & \cgreen 0 & \cgreen1 & \cred1 & \cred1 & \cgreen0 & \cred1 \\
  \tau_\pi & \cred1 & \cred1 & \cgreen0 & \cgreen1 & \cred1 & \cgreen0 \\ \hline
\end{array}
$$
It is easy to see that
$\tau(\eqref{eq:example})=\tau_{\pi}(\eqref{eq:example}))=0$. This is
not a coincidence, it is actually a property that derives from the
shape of $\Phi(X)$. This is stated in the following property.
\end{example}
\end{arxiv}
\begin{proposition} 
\label{prp:permutation-invariance-truth-assignment}
For every pure universal formula $\Phi(\bx)$, every permutation $\pi$
of $X$ and  every lifted interpretation $\tau$ for $\Phi(X)$, 
$\tau(\Phi(X)) = \tau_{\pi}(\Phi(X))$;
where $\tau_\pi(P(x_i,x_j,\dots)=
\tau(P(\pi(x_i),\pi(x_j),\dots)$, for every atom $P(x,y,\dots)$.
\end{proposition}

\begin{proof}
If $\tau(\Phi(X))=0$ then $\tau(\Phi(\bx'))=0$ for some
$\bx'\in X^m$. This implies that $\tau_\pi(\Phi(\pi^{-1}(\bx')))=0$,
which implies that $\tau_\pi(\Phi(X))=0$. The proof
of the opposite direction follows form the fact that
$(\tau_\pi)_{\pi^{-1}}=\tau$. 
\end{proof}

From now on, we concentrate on the special case where $X=\{x,y\}$ i.e. FO$^2$.
A closed-form formula for \fomc\ in FO$^2$ has been proved in
  \cite{Symmetric_Weighted}. In the following we reconstruct this result
  using the notion of lifted interpretations. As it will be clearer
  later, using lifted interpretation allows us to
  seamlessly extend the closed-form to larger extensions of FO$^2$
  formulas.

For any lifted interpretation $\tau$ of $\Phi(X)$, let
$\tau_x$ and $\tau_y$ be the partial lifted interpretation that assign only the atoms
containing $x$ and $y$ respectively.
Notice that if $P(x)$ is an atom of $\Phi(X)$, so is $P(y)$ and
vice-versa. This implies that $\tau_x$ and $\tau_y$ assign two sets
of atoms that are isomorphic under the exchange of $x$ with $y$. 
Let $u$ be the number of atoms contained in each of these two sets
and let $P_0,\dots,P_{u-1}$ be an enumeration of the predicate symbols of
these atoms. In other words, we have $\tau_x$ that assigns truth value to 
$P_0(x),\dots,P_{u-1}(x)$ and $\tau_y$ that assigns to
$P_0(y),\dots,P_{u-1}(y)$.%
\footnote{When the atoms are $P_j(x,x)$ or $P_j(y,y)$, i.e, when 
 $P_j$ is a binary predicate, 
 with an abuse of notation, we denote these atoms with $P_j(x)$ and $P_j(y)$.}
This implies that $\tau_x$ and $\tau_y$ 
can be represented by two integers $i$ and $j$ respectively between
$0$ and $2^u-1$,
such that $\tau_x = i$ if and only if $\tau_x(P_k(x))=bin(i)_k$, and
$\tau_y = j$ if and only if $\tau_y(P_k(y))=bin(j)_k$, where $bin(i)_k$ refers to the $k^{th}$ number (0 or 1) of the binary encoding of the integer $i$.
For every $0\leq i,j\leq 2^u-1$, we define $n_{ij}$ as the number of
lifted interpretations of $\Phi(X)$ which are extensions of the partial lifted interpretations $\tau_x=i$ and
$\tau_y=j$. Hence, $n_{ij}$ can be written as follows (where we
consider variables as constants)
$$
n_{ij}=\mc(\Phi(X)\wedge
\bigwedge_{k=0}^{u-1}\left(\neg^{1-bin(i)_k}P_k(x) \wedge \neg^{1-bin(j)_k}P_k(y)\right)
$$
where $\neg^0$ is the empty string and $\neg^1$ is $\neg$.
Notice that Proposition~\ref{prp:permutation-invariance-truth-assignment}
guarantees that $n_{ij}=n_{ji}$.

\begin{example}[Example \ref{ex:running} cont'd]
  \label{ex:n-ij}
  The set of atoms containing only $x$ or only $y$ in the formula
  \eqref{eq:example} are $\{A(x), R(x,x)\}$ and $\{A(y), R(y,y)\}$
  respectively. In this case $u=2$. The partial lifted interpretations
  $\tau_x$ and $\tau_y$ corresponding to the lifted
    interpretation $\tau$
  of Example~\ref{ex:truth-assignment} are:
  $\tau_x=1$ and $\tau_y=3$. $n_{13}$ is the number of lifted interpretations
  satisfying   \eqref{eq:example} and agreeing with
  $\tau_x=1$ and $\tau_y=3$. In this case $n_{13}=2$.
  The other cases are as follows:
  $$
 \begin{array}{cccccccccc} \hline 
   n_{00} &    n_{01} &    n_{02} &    n_{03} &    n_{11} &    n_{12}
   &    n_{13} &    n_{22} &    n_{23} &    n_{33} \\
   4 &    4 &    2 &    2 &    4 &    2 &
                                                                  2
               &    4 &    4 &    4 \\ \hline 
 \end{array}
 $$
\end{example}

\begin{arxiv}
For any set of constants $C$ and any $2^u$-tuple $\bk=(k_0,\dots,k_{2^u-1})$
such that $\sum\bk=\card{C}$, let $\CC_\bk$ be any partition $(C_i)_{i=1}^{2^u-1}$
of $C$ such that $\card{C_i}=k_i$. 
We define  $\Phi(\mathbb{C}_{\bm{k}})$ as follows: 
\begin{equation}
  \label{eq:phi-c-k}
    \Phi(\CC_\bk) = \Phi(C) \wedge
\bigwedge_{i=0}^{2^u-1}\bigwedge_{c\in C_i}
\bigwedge_{j=0}^{u-1}(\neg)^{1-bin(i)_j}P_j(c)
\end{equation}

\begin{example}
Examples of $\CC_{(1,0,2,0)}$, on $C=\{a,b,c\}$ are 
$\{\{a\},\emptyset,\{b,c\},\emptyset\}$ and
$\{\{b\},\emptyset,\{a,c\},\emptyset\}$. 
\begin{align*}
    \Phi(\{ \{a\},\emptyset,\{b,c\},\emptyset\})  = \Phi(C) &\land \neg A(a) \land \neg R(a,a)\\
    &\land  A(b) \land \neg R(b,b)\\
    &\land A(c) \land \neg R(c,c)  \\
\end{align*}
Note there are ${3 \choose 1,0,2,0}=3$ such partitions, and all the $\Phi(\CC_\bk)$ for such partitions will have the same model count. These observations have been formalized in lemma \ref{lem:mc-Phi-C}
\end{example}

\begin{lemma}
\label{lem:mc-Phi-C}
$    \mc(\Phi(C)) = \sum_{\bk}\binom{n}{\bk} \mc(\Phi(\mathbb{C}_{\bm{k}}))$
\end{lemma}
\begin{proof}
  Let $\CC_\bk$ and $\CC'_\bk$, be two partitions with the same $\bk$.
  Notice that $\CC'_{\bk}$ can be obtained by applying some
  permutation on $C$ from $\CC_\bk$. From
  Proposition~\ref{prp:permutation-invariance-truth-assignment} 
  we have that
  $$\mc(\Phi(\CC_\bk))=\mc(\Phi(\CC'_\bk))$$
  Furthermore notice that if $\CC_\bk$ is different from $\CC'_{\bk'}$
  then $\Phi(\CC_\bk)$ and $\Phi(\CC'_{\bk'})$ cannot be
  simultaneously satisfied. This implies that
  $$
  \mc(\Phi(C)) = \sum_{\bk}\sum_{\CC_\bk}\mc(\Phi(\CC_\bk))
  $$
  Since there are ${n \choose \bk}$ partitions of $C$, of the
  form $\CC_\bk$, then 
  $$
  \mc(\Phi(C)) = \sum_{\bk}{n\choose\bk}\mc(\Phi(\CC_\bk))
  $$
\end{proof}

\begin{lemma}
\label{lem:join}
For any partition $\CC_\bk=\{C_0,\dots,C_{2^u-1})$ 
$$\mc(\Phi(\CC_\bk)) = \prod_{\substack{c\neq d  \\ c,d \in C}} n_{i_c i_d}$$ 
Where  for all $c,d\in C$, $0\leq i_c,i_d\leq 2^u-1$ are the indices such that 
$c\in C_{i_c}$ and $d\in C_{i_d}$. 
\end{lemma}

\begin{proof}
  $\Phi(\CC_\bk)$ can be rewritten in
  $$
  \bigwedge_{\{c,d\}\subseteq C \atop c\neq d}\Phi^{i_c,i_d}(\{c,d\})
  $$
  $\Phi^{i_c,i_d}(\{c,d\}))$ is obtained by replacing each atom
  $P_j(c)$ with $\top$ if $bin(i_c)_j=1$ and $\bot$ otherwise
  and each atom $P_j(d)$ with $\top$ if
  $bin(i_d)_j=1$ and $\bot$ otherwise.
  Notice that all the atoms of 
  $\Phi^{i_c,i_d}(\{c,d\})$ contain both $c$ and $d$.
  Furthermore notice that if $\{c,d\}\neq\{e,f\}$ then
  $\Phi^{i_c,i_d}(\{c,d\}))$ and $\Phi^{i_e,i_f}(\{e,f\}))$ do not
  contain common atoms.
  Finally we have that $\mc(\Phi^{i_c,i_d}(\{c,d\}))=n_{i_ci_d}$.
  Hence
  $$
  \mc\left(\bigwedge_{c, d\in C\atop c\neq
      d}\Phi^{i_c,i_d}(\{c,d\})\right)= \prod_{\substack{c\neq d  \\
      c,d \in C}} n_{i_c i_d}
  $$ 
\end{proof}
\end{arxiv}
\begin{theorem}
  \label{thm:mc-pure-univ-formulas}
  For any pure universal formula \footnote{Our results for the pure universal formula are similar to \cite{Symmetric_Weighted}, with substantial change in notation.}
 \begin{align}
    \label{eq:fomc-universal-form}
  \fomc(\forall \bx. \Phi(\bx),n)=\sum_{\sum \bk=n}  {n \choose \bm{k}}
  \prod_{0 \leq i \leq j \leq 2^u-1} n_{i j}^{\bk(i,j)}
\end{align}
\begin{align}
  \bk(i,j) =
  \begin{cases}
      \frac{k_{i}(k_{i} - 1)}{2} & \text{if $i=j$} \\
       k_{i}k_{j} & \text{otherwise} \\
     \end{cases}
\end{align}
\end{theorem}
Notice, theorem \ref{thm:mc-pure-univ-formulas} deals with equality implicitly in the lifted interpretations, which requires constant time w.r.t domain cardinality. 
\begin{proof} 
  Notice that $\fomc(\phi(\bx),n)=\mc(\Phi(C))$ for a set of constants
  $C$ with $\card{C}\ =n$. Therefore, by Lemma~\ref{lem:mc-Phi-C}, to prove
  the theorem it is enough to show that for all $\bk$,
  $\mc(\Phi(\CC_{\bk}))= \prod_{0 \leq i \leq j \leq 2^u-1}
  n_{ij}^{\bk(i,j)}$. By the Lemma~\ref{lem:join} we have that 
$ \mc(\Phi(\CC_\bk)) = \prod_{c\neq d} n_{i_c i_d}$. Then:
\begin{align*}
\prod_{c\neq d} n_{i_c i_d} & = 
  \prod_{i} \prod_{\substack{c \neq d \\  c,d \in C_{i}}} n_{ii} \cdot \prod_{i< j}\prod_{\substack{c \in C_i \\ d \in C_j }} n_{ij}  \\
 & =  \prod_{i} n_{ii}^{k_{i} \choose 2} \cdot \prod_{i < j}  n_{ij}^{k_{i}k_{j}} =
 \prod_{0 \leq i \leq j < 2^u} n_{ij}^{\bk(i,j)}
\end{align*}
\end{proof}

\begin{example}[Example \ref{ex:running} cont'd]
  \label{ex:cardinality}
  Consider a domain of 3 elements (i.e., n=3). Each term of the summation
  \eqref{eq:fomc-universal-form} is of the form
  $$
  \binom{3}{k_0,k_1,k_2,k_3}\prod_{0\leq i\leq j < 2^u-1}n_{ij}^{\bk(i,j)}
  $$
  which is the number of models with $k_0$ elements for which $A(x)$
  and $R(x,x)$ are both false; $k_1$ elements for which $A(x)$ is false
  and $R(x,x)$ true, $k_2$ elements for which $A(x)$ is true and
  $R(x,x)$ is false and $k_3$ elements for which $A(x)$ and
  $R(x,x)$ are both true. For instance
  $$
  \binom{3}{2,0,0,1} n_{00}^1n_{03}^2= \binom{3}{2,0,0,1} 4^1\cdot 2^2
  = 3\cdot 16 = 48
  $$
  is the number of models in which 2 elements are such that $A(x)$ and
  $R(x,x)$ are false and $1$ element such that $A(x)$ and $R(x,x)$ are both
  true. 
  \end{example}

\begin{arxiv}
  As a final remark for this section, notice that the computational cost of computing $n_{ij}$ is constant with respect to the domain cardinality. We assume the cost of multiplication to be constant. Hence, the computational complexity of computing  \eqref{eq:fomc-universal-form} depends on the domain only through the multinomial coefficients $\binom{n}{\bk}$ and the multiplications involved in $\prod_{ij} n^{\bk(i,j)}$. The computational cost of computing  $\binom{n}{\bk}$ is polynomial in $n$ and the total number of $\binom{n}{\bk}$  are $\binom{n+2^{u}-1}{2^{u}-1}$, which has
  $\left(\frac{e\cdot(n+2^{u}-1)}{2^{u}-1}\right)^{2^{u}-1}$ as an upper-bound\cite{das2016brief}. Also, the $\prod_{ij} n^{\bk(i,j)}$ term has  $O(n^2)$ multiplication operations. Hence, we can conclude that the \eqref{eq:fomc-universal-form} is computable in polynomial time with respect to the domain cardinality. 
\end{arxiv}


\section{FOMC for Cardinality Constraints}
Cardinality constraints are arithmetic constraints on the number of true interpretations of a set of predicates in a given FOL formula. 
In Example~\ref{ex:cardinality}, we showed how different values of $\bk$ can represent different unary predicate cardinalities. 
Let's formalize the correspondence between the multinomial factor $n \choose \bk$ and the cardinality of
the unary predicates \begin{arxiv} of the models that satisfy $\Phi(\CC_\bk)$ \end{arxiv}.  For
every $\bk$ with $\sum\bk=n$ and for every unary predicate $P_j$,
we define
$$
\bk(P_j)=\sum_{0\leq i \leq 2^u-1}bin(i)_j\cdot k_i
$$
The following lemma states that $\bk(P_j)$ is the number of $c\in C$ such that  $\omega(P_j(c))=1$. 
\begin{lemma}
For every $2^u$-tuple of non-negative integers $\bk$ with $\sum\bk=n$, and every
unary predicate $P_j$, and every truth assignment $\omega$, if ${\omega}\models\Phi(\CC_\bk)$ then
$\sum_{c\in C}\omega(P_j(c))=\bk(P_j)$.\begin{ijcai}Where, $\CC_\bk$ is a partition of the domain such the truth assignment to unary predicate agrees with $\bk$. \end{ijcai}
\end{lemma}
\begin{proof}
  The lemma follows immediately from the definition of $\Phi(\CC_{\bk})$ given in
  equation \eqref{eq:phi-c-k}.
\end{proof}

Let  $\rho(\{P_i\})$ be any arithmetic constraint 
on the integer variables representing the cardinality of unary
predicates in the $n$-tuple $\{P_i\}$. We say that 
$\bk\models\rho(\{P_i\})$, 
if $\rho$ is satisfied when each integer variable, representing cardinality of $P_i$, is substituted for 
the integer $\bk(P_{i})$ in $\rho$.
\begin{corollary}[of Theorem~\ref{thm:mc-pure-univ-formulas}]
\label{cor:mc-pure-univ-formulas-card-constr-unary}
  For every cardinality restriction $\rho$ on unary predicates, 
  \begin{align}
   \label{eq:fomc-univ-card-constr}
\fomc(\forall \bx\Phi(\bx)\land \rho, n) & = 
\sum_{\bk\models \rho}  {n \choose \bm{k}}
  \prod_{0 \leq i \leq j \leq 2^u-1}\!\!\!n_{i j}^{\bk(i,j)}\!\!\!
  \end{align}
\end{corollary}
\begin{example}
To count the models of \eqref{eq:example} with
the additional constraint that $A$ is balanced i.e.,
\mbox{$\frac{n}{2}\leq\card{A}\leq\frac{n+1}{2}$}, we have to consider only the
terms where $\bk$ is such \mbox{$\frac{n}{2}\leq
  \bk(A)\leq\frac{n+1}{2}$}.
Equivalently in equation \eqref{eq:fomc-univ-card-constr} we should
consider only the $\bk$ such that $\frac{n}{2}\leq k_2+k_3\leq
\frac{n+1}{2}$. (Notice that $k_2$ is the number of elements that
satisfy $A(x)$ and $\neg R(x,x)$ and $k_3$ is the number of elements
that satisfy $A(x)$ and $R(x,x)$). 
\end{example}
To count models that satisfy cardinality restriction on
binary predicates, we need to extend the result of
Theorem~\ref{thm:mc-pure-univ-formulas}.
Similar to what we have done for unary atoms, let
$R_0(x,y),R_1(x,y),\dots,R_b(x,y)$ be an enumeration of the atoms of
$\Phi(X)$ that contain both variables $x$ and $y$. Notice that the
order of variables accounts towards different predicates, for instance in Example
\ref{ex:running},  we have two predicates $R_1(x,y)=R(x,y)$ and
$R_2(x,y)=R(y,x)$. Every assignment of a lifted interpretation to these predicates can be
represented with an integer $v$, with $0\leq v \leq 2^b-1$, with the
usual convention that, if $\tau_{xy}=v$, then
$\tau_{xy}(R_k(x,y))=bin(v)_k$. Now for every $1\leq i \leq j \leq
2^u-1$ and every $0\leq v\leq 2^b-1$,
$n_{ijv}=\tau_{x}\tau_y\tau_{xy}(\Phi(X))$,
where $\tau_x=i$, $\tau_y=j$ and $\tau_{xy}=v$. 
We start by observing that
\begin{align}
   \label{eq:n-ijuv-expansion}
\mbox{$n_{ij}=\sum_{v=0}^{2^{b}-1}n_{ijv}$}
\end{align}
\begin{example}
  For instance $n_{13}$ introduced in Example~\ref{ex:n-ij}
  expands to $n_{130}+n_{131}+n_{132}+n_{133}$
  where $n_{13v}$ corresponds to the following assignments:
  $$
\arraycolsep=2pt
  \begin{array}{|cc;{1pt/1pt}cc;{1pt/1pt}cc|cc|} \hline 
 A(x) & R(x,x) & A(y) & R(y,y) & R(x,y) & R(y,x) & v & n_{13v}\\ \hline
    \multirow{4}{*}{$0$} & 
    \multirow{4}{*}{$1$} & 
    \multirow{4}{*}{$1$} & 
    \multirow{4}{*}{$1$} & 
    0 & 0 & 0 & n_{130} = 1 \\ 
    & & & & 0 & 1 & 1 & n_{131}=0\\ 
    & & & & 1 & 0 & 2 & n_{132}=1\\ 
    & & & & 1 & 1 & 3 & n_{133}=0\\ \hline 
    \multicolumn{2}{c}{\tau_x=1} 
       & \multicolumn{2}{c}{\tau_y=3}
   & \multicolumn{2}{c}{\tau_{xy}=v} \\ 
  \end{array}
$$

\end{example}
\begin{arxiv}
Notice that $n_{ijv}$ is either $0$
or $1$. 
\end{arxiv}
By replacing $n_{ij}$ in equation \eqref{eq:fomc-universal-form} with
with its expansion \eqref{eq:n-ijuv-expansion} we obtain that
$\fomc(\Phi(\bx),n)$ is equal to 

  \begin{align}
    \nonumber 
& 
  \sum_{\sum\bk=n}  {n \choose \bm{k}}
  \prod_{0 \leq i \leq j \leq 2^u-1} \left(\sum_{0\leq v \leq
                        2^b-1}n_{i j v}\right)^{\bk(i,j)} \\
    \nonumber
                   & = \sum_{\bk,\bh}{n \choose \bk}
\prod_{0 \leq i \leq j \leq 2^u-1}
{\bk(i,j) \choose \bh^{ij}}
\prod_{0\leq v\leq 2^b-1} 
    n_{ijv}^{h^{ij}_{v}} \\ 
   \label{eq:fomc-universal-form-expanded}
& = \sum_{\kh}F(\kh,\{n_{ijv}\})  
\end{align}
where, for every $0\leq i\leq j\leq 2^u-1$, $\bh^{ij}$ is a vector of
$2^b$ integers that sum up to $\bk(i,j)$, and in
\eqref{eq:fomc-universal-form-expanded}, to simplify the notation, we define the term in the summation
corresponding to $\kh$ as $F(\kh,\{n_{ijv}\})$

Similarly to what we have done for unary predicates, we define $\bh_{ij}(R)$
for every binary predicate $R$ as follows: 
\begin{align}
  \bh_{ij}(R) & =
\sum_{v=0}^{2^b-1}(bin(v)_l+bin(v)_r)\cdot h^{ij}_{v}
\end{align}
where $l$ and $r$ are the indices such that $R_l$ corresponds to
$R(x,y)$ and $R_r$ to $R(y,x)$. For every predicate $P$ we define
$(\kh)(P)$ as $\bk(P)$ if $P$ is unary and $\bk(P) + \bh(P)$ if $P$ is
binary. For an $n$-tuple of predicates $\{P_i\}$, we use $(\kh)(\{P_i\})$ to 
denote the $n$-tuple of non-negative integers $\{(\kh)(P_i)\}$. 
\begin{example}
A graphical representation of the pair $\bk,\bh$ for the formula
\eqref{eq:example} is provided in the following picture: 
\begin{center}
  \begin{tikzpicture}
    \foreach \i in {0,...,3}{
      \node at (1.2*\i cm,0.75) {$k_\i$};
      \node at (-1,-\i cm) {$k_\i$};      
       \foreach \j in {\i,...,3}
        \node[rectangle,draw,minimum width=1.2cm, minimum height =
        1cm,inner sep = 1pt ]
        at (1.2*\j cm,-\i cm) (h\i\j)
        {$\begin{smallmatrix}h^{\i\j}_0 & h^{\i\j}_1 \\ h^{\i\j}_2 &  h^{\i\j}_3\end{smallmatrix}$};};
    \end{tikzpicture}
  \end{center}
This configuration represent the models in which a set $C$ of $n$ 
constants are partitioned in four sets $C_0,\dots,C_3$, each $C_i$ containing
$k_i$ elements (hence $\sum k_i=n$). Furthermore, for each pair $C_i$ and $C_j$ the relation
$D^{ij}=C_i\times C_j$ is  partitioned in 4 sub relations
$D^{ij}_0,\dots,D^{ij}_3$ where each $D^{ij}_v$ contains $h^{ij}_v$
pairs (hence $\sum_vh_v^{ij}=\bk(i,j)$). For instance if the pair $(c,d)\in D^{12}_2$ it means that
we are considering assignments that satisfy
$\neg A(c) \land R(c,c) \land A(d) \land \neg R(d,d) \land R(c,d)
\land \neg R(d,c)$.
\end{example}
Let  $\rho(\{P_{i}\})$ be any arithmetic constraint 
on the integer variables representing the cardinality of the set of
predicates $\{P_i\}$. We write $(\bk,\bh)\models\rho(\{P_i\})$ to denote that
the cardinality constraint $\rho((\kh)\{P_i\})$ is satisfied. 

\begin{corollary}[of Theorem~\ref{thm:mc-pure-univ-formulas}]
  For every cardinality restriction
  $\rho(\{P_i\})$, and every pure universal formula $\Phi(\bx)$, 
$\fomc(\forall \bx\Phi(\bx)\land \rho(\{P_i\}), n)    
    =\sum_{\kh\models\rho}F(\kh,\{n_{ijv}\})
$
\end{corollary}

\begin{example}
  Consider formula \eqref{eq:example} with the additional conjunct
  $\card{A}= 2$ and $\card{R}=2$. The constraint $\card{A}=2$ implies that we have to
  consider $\bk$ such that $k_2+k_3=2$. $\card{R}=2$ constraint translates to only
  considering monomials with
  $k_1+k_3+h^{ij}_{1}+ h^{ij}_{2} + h^{ij}_{3} = 2$. 
\end{example}

\section{FOMC for Existential Quantifiers}
Any arbitrary formula in FO$^2$ can be reduced to an equisatisfiable
reduction called Scott's Normal Form(SNF) \cite{Scott1962},
\begin{arxiv}see equation \eqref{eq:scott-form}\end{arxiv}
\begin{ijcai}see \eqref{eq:simple-scott} for an example of SNF\end{ijcai}.
\cite{kuusisto2018weighted} prove
that SNF also preserves WFOMC of the FO$^2$ formulas. In this section,
we reconstruct the result given in \cite{Symmetric_Weighted} by
extending our result for FOMC in universally quantified formulas to
the whole FO$^2$ fragment by providing an FOMC formula for SNF. 
The main difference w.r.t. \cite{Symmetric_Weighted} is that we
  explicitly use the inclusion and exclusion principle, instead of
  introducing negative weights.
We first consider the following simpler case:
    \begin{align}
      \label{eq:simple-scott}
    \forall x \forall y.\Phi(x,y) \land \forall x \exists y.\Psi(x,y)
    \end{align}
    where $\Phi(x,y)$ and $\Psi(x,y)$ are formulae without quantifiers.
    First of all notice that: 
    \begin{align}
      \label{eq:difference}
      \fomc(\eqref{eq:simple-scott},n) & =
      \fomc(\forall x y.\Phi(x,y),n) \\ & - 
      \fomc(\forall xy.\Phi(x,y)\wedge\exists x\forall y\neg\Psi(x,y),n)
      \nonumber 
    \end{align}
    The first term of \eqref{eq:difference} can be computed by
    Theorem~\ref{thm:mc-pure-univ-formulas}; for the second term we 
    need to prove an auxiliary lemma, which uses the following
    notation: 
    \def\PhiE#1{\mbox{$\Phi(x,y) \land\exists^{=#1}x\forall y\neg\Psi(x,y)$}}
    \def\PhiP#1{\mbox{$\Phi(x,y)\land (P(x)\rightarrow\neg \Psi(x,y))\land \card{P}=#1$}}
    \begin{align*}
    e_m & = \fomc(\forall xy.\PhiE{m},n) \\
    p_m & = \fomc(\forall xy.\PhiP{m},n)
    \end{align*}
    where $P$ is a new unary predicate. In the following lemma we show
    that $e_m$ can be expressed as a function of $p_i$'s. 
    
    \begin{lemma}
    \label{lem:E-m}
    $$
    e_m = \sum_{k=m}^{n} (-1)^{k-m}\binom{k}{m}p_{k}
    $$
    \end{lemma}
    
    \begin{proof}[Proof of lemma~\ref{lem:E-m}]
    By induction on $m-n$
    \paragraph{\underline{$m=n$}} The lemma holds since 
    $\PhiE{n}$ is equivalent to \\ $\PhiP{n}$ when the domain cardinality is
    $n$. 
    \paragraph{\underline{$m+1 \implies m$}}
    \begin{align}
    e_m & = p_m - \sum_{k=m+1}^n\binom{k}{m} e_{k} \\
    \label{prf:before-induction}
    & \stackrel{ind}{=} 
    p_m - \sum_{k=m+1}^n\binom{k}{m} 
    \sum_{h=k}^{n} (-1)^{h-k}\binom{h}{k}p_{h} \\
    \label{prf:after-induction}
    & = \sum_{k=m}^{n} (-1)^{k-m}\binom{k}{m}p_{k}
    \end{align}
    The equality of \eqref{prf:before-induction} and
    \eqref{prf:after-induction} can be obtained by expanding
    the summation and showing that all the terms of every internal
    summation cancel but one. We omit this expansion since it is
    routinary. 
    \end{proof} 
    
    \begin{example}
    An expansion of the statement of Lemma \ref{lem:E-m} with $m=3$ and
    $n=4$ is
    $e_2  = {2 \choose 2}p_2 - {3 \choose 2} p_3 + {4 \choose 2} p_4$
    \end{example}
    Since $p_m$ is the first order model count of a pure universal
    formula with cardinality restriction, it can be computed by the formula of
    Corollary
    \ref{cor:mc-pure-univ-formulas-card-constr-unary}. Lemma~\ref{lem:E-m}
    tells us how to compute also $e_m$ starting from the $p_m$'s.
    Finally notice that, the second term of equation
    ~\eqref{eq:difference} can be computed by summing $e_m$ from $1\leq m
    \leq n$. This is possible since the set of models counted in $e_m$ are
    disjoint from the set of models counted in
    $e_{m'}$, where $m\neq m'$.  This allows us to state the following
    theorem: 
    \begin{theorem}
    \label{thm:fomc-simple-scott}
    Let $\Phi'(x,y)$ be the formula $\Phi(x,y)\land (P(x)\rightarrow
    {\neg}\Psi(x,y))$ and let $n_{ij}$ be the number of lifted interpretations of 
    $\Phi'(X)$ which are extensions of the partial lifted
      interpretation $\tau_x=i$ and $\tau_y=j$,
    then 
    \begin{align}
    \fomc(\eqref{eq:simple-scott},n) = 
      \sum_{\sum \bk=n}  {n \choose \bm{k}}(-1)^{\bk(P)}\hspace{-1em}
      \prod_{0 \leq i \leq j \leq 2^u-1} n_{i j}^{\bk(i,j)}
    \end{align}
    \end{theorem}
    
    \begin{proof}
      \begin{align*}
    \fomc(\eqref{eq:simple-scott},n) & = p_0 - \sum_{m=1}^n e_m  \\
    &\text{by Lemma \ref{lem:E-m}} \\
    & = p_0 - \sum_{m=1}^n\sum_{k=m}^n(-1)^{k-m}{k \choose m}p_k \\
    & = p_0 - \sum_{k=1}^{n}(-1)^{k+1}p_k  \\
    & = \sum_{k=0}^{n}(-1)^{k}p_k
      \end{align*}
  Which is the same as the equation proposed in the theorem. Hence, completing the proof. 
        \end{proof}
    


    We generalize the previous result to compute first order
    model counting for FO$^2$ formulas in Scott's normal form. 
    \begin{theorem}
    \label{thm:fomc-scott-form}
    Consider a formula in Scott's normal form
    \begin{align}
    \label{eq:scott-form}
    \forall x y.\Phi(x,y)\land\bigwedge_{i=1}^q\forall x\exists y.\Psi_i(x,y)
    \end{align}
    Where, $\Phi(x,y)$ and $\Psi_i(x,y)$ are quantifier free
    formulas. Let $\Phi'(x,y)$ be the formula
    $\Phi(x,y)\land \bigwedge_{i=1}^{q}(P_i(x)\rightarrow
    \neg\Psi_i(x,y))$, where $P_i$'s are fresh unary
    predicates, let $n_{ij}$ be the number of lifted interpretations of $\Phi'(X)$
    which are extensions of the partial lifted interpretation $\tau_x=i$ and
    $\tau_y=j$, then
    \begin{align}
    \label{eq:fomc-scott-form}
    \fomc(\eqref{eq:scott-form},n) = \!\!
      \sum_{\sum \bk=n}  {n \choose \bm{k}}(-1)^{\sum_l\bk(P_l)}\hspace{-1.5em}
      \prod_{0 \leq i \leq j \leq 2^u-1} n_{i j}^{\bk(i,j)}
    \end{align}
    \end{theorem}
    
    \begin{proof}[outline]
      We generalize Lemma~\ref{lem:E-m} as follows:\\
      For every $\bm m= (m_1,\dots,m_q)$ with $0\leq m_i\leq n$ we define 
      \begin{align*}
    e_{\bm m} & = \fomc(\forall xy
                \Phi(x,y)\wedge\bigwedge_{i=1}^{q}\forall x
                \exists^{=m_i}y\neg\Psi_i(x,y),n) \\
    p_{\bm m} & =\\
    &\fomc(\forall xy
                \Phi(x,y)\wedge\bigwedge_{i=1}^{q}P_i(x)\rightarrow\neg\Psi_i(x,y)
                \wedge \card{P_i}\ = m_i,n) \\
      \end{align*}
    The proof of Lemma ~\ref{lem:E-m} can be generalized to show
    that:
    \begin{align}\nonumber
      e_{\bm m} & = \\
      &\sum_{k_1=m_1}^n(-1)^{k_1-m_1}{k_1 \choose m_1}\dots 
                  \sum_{k_q=m_q}^n(-1)^{k_q-m_q}{k_q \choose
                  m_q}p_{k_1,\dots,k_q} \\
      \label{eq:e-mm}
                & = \sum_{k_1=m_1}^n\dots\sum_{k_q=m_q}^n(-1)^{\sum_{i=1}^q
        k_i-m_i}p_{k_1,\dots,k_q}\prod_{i=1}^q{k_i\choose m_i}
    \end{align}
    Using a generalization of equation \eqref{eq:difference} we have that: 
    \begin{align*}
      \fomc(\eqref{eq:scott-form},n) & = p_{0,\dots,0} - \sum_{
                                       \sum\bm m \geq 1}^{(n,\dots,n)}e_{\bm
                                       m}
    \end{align*}
    The proof of \eqref{eq:fomc-scott-form} can be obtained by replacing
    the $e_{\bm m}$ with equation \eqref{eq:e-mm} and
    simplifying as in the proof of Theorem~\ref{thm:fomc-simple-scott}.
    \end{proof}

    As a final remark, notice that FOMC for FO$^2$ formulas with
    cardinality on unary and binary predicate can be computed
    by first expanding \eqref{eq:fomc-scott-form} in order to take into account also
    $\bh$, and then restricting to the $(\bk,\bh)$ that satisfy
    $\rho$. We, therefore, obtain that for an FO$^2$ formula $\Phi$ in Scott Normal
    Form $\fomc(\Phi\wedge\rho,n)$ is equal to 
    \begin{align}
    \label{eq:fomc-scott-form-kh}
    \sum_{\bk,\bh\models\rho}  {n \choose \bm{k}}(-1)^{\sum_l\bk(P_l)}\hspace{-1.8em}
      \prod_{0 \leq i \leq j \leq 2^u-1}{\bk(i,j)\choose
      \bh^{ij}}\prod_{0\leq v\leq 2^b-1}\!\!\!n_{ijv}^{h^{ij}_v}
    \end{align}
  


\def\bc{\bm{c}}
\def\wfomc{\text{\footnotesize\sc wfomc}}
\section{Weighted First Order Model Counting}
In FOMC every model of a formula contributes with one unit
to the final result. Instead in WFOMC, models
can be associated with different contributions, also called
\emph{weights}. The weight of an interpretation $\omega$ is 
provided by a weight function $w$
that associates a real number to it. More formally: given
a first order language $\L$ and an interpretation domain $C$
a weight function $w$ is a function $w:\omega\mapsto w(\omega)\in\R$.
WFOMC has been extensively studied for finite domains, and for weight
functions that are independent of individual domain elements.
In this case the definition of weighted model counting reduces to 
$
\wfomc(\Phi,w,n) = \sum_{\omega\models\Phi}w(\omega)
$
where $n$ is the cardinality of the domain. We propose a new family 
of such weight functions on $(\kh)$ vectors. A weight function $w(\kh)$ associates 
a real number to each $(\kh$). 
Hence, we define WFOMC  as follows :
\begin{definition}
For all $\Phi$ in FO$^2$ and for arbitrary cardinality 
constraint
$\rho$.

$$
\wfomc(\Phi,w,n) = \!\!\!\!\sum_{\bk,\bh\models \rho}\!\!\!\!w(\bk,\bh)\cdot
F\left(\bk,\bh,\{n_{ijv}\}\right)
$$
where $w(\bk,\bh)$ is an arbitrary positive real valued function. 
\end{definition}

\subsection{Symmetric Weight Functions}
\emph{Symmetric weight function} \cite{Symmetric_Weighted} is a
family of weight functions that  can be specified by
a function $w:\preds\times \{0,1\}\rightarrow\R$, where $\preds$ is the 
set of predicate symbols of $\L$. The weight of an assignment
$\omega$ is then defined as follows: 
$$
w(\omega)=\prod_{P(\bm c)\in atoms(\L)} w(P,\omega(P(\bc)))
$$
The following theorem shows how symmetric weight functions can be 
expressed by $w(\kh)$. 
\begin{theorem}
For all $\Phi$ in FO$^2$ and for arbitrary cardinality 
constraint
$\rho$, symmetric-WFOMC
can be obtained from FOMC by defining the following weight function:
\begin{align*}
w(\bk,\bh) & = \prod_{P\in\L}
w(P,1)^{(\bk,\bh)(P)}\cdot 
w(P,0)^{(\bk,\bh)(\neg P)}
\end{align*}
where $(\kh)(\neg P)=n-\bk(P)$ if $P$ is unary and $n^2-(\kh)(P)$ if $P$
is binary.
\end{theorem}

\begin{proof}
The proof is a consequence of the observation that
$F(\bk,\bh,\{n_{ijv}\})$ is the number of models of $\Phi$ that
contains $\bk(P)$ elements that satisfies $P$, if $P$ is unary,
and $(\bk,\bh)(P)$ pairs of elements that satisfy $P$, if $P$ is
binary. 
\end{proof}

\subsection{Expressing Count Distributions}
Symmetric weight functions cannot express many interesting
distributions. 
For instance, consider a set of domain elements $C=\{c_1,\dots,c_n\}$ which have
an attribute $A$. To impose a fairness constraint on $A$, 
one would like to have higher weights for interpretations $\omega$ in which
the number instances of $A$ being true and false are balanced, e.g., 
it is proportional to $\left(\card{A^\omega}-\card{\neg A^\omega}\right)^2$.
A class of weight functions that allow modelling these type of situations
have been introduced in \cite{complexMLNkuzelka2020}. These weight functions 
have been introduced to express
count distributions, which are defined in the following definition.

\begin{definition}[Count distribution \cite{complexMLNkuzelka2020}]
Let $\Phi = \{\alpha_i, w_i\}_{i=1}^m$
be a Markov Logic Network defining a distribution over a set of
possible worlds (we call them assignments)
$\Omega$. The count distribution of $\Phi$ is the distribution over
$m$-dimensional vectors of non-negative integers $\bm n$ given by
\begin{align}
q_{\Phi}(\Omega,\bm n) & = \sum_{\omega\in\Omega,\ \bm n =
                   \bm N(\Phi,\omega)} p_{\Phi,\Omega}(\omega)
\end{align}
where $\bm N(\Phi,\omega)=(n_1,\dots,n_m)$, and $n_i$ is the number of
grounding of $\alpha_i$ that are true in $\omega$.
\end{definition}

\cite{complexMLNkuzelka2020} shows that count distributions can
be modelled by MLN's with complex weights. In the following, we show 
that if $\alpha_i$ and $\Phi$ are in FO$^2$, then we can express
count distributions with positive real valued weights on $(\kh)$. 

\begin{theorem}
Every count distribution over a set of possible worlds $\Omega$ 
definable in FO$^2$ can be modelled with a weight
function on $(\bk,\bh)$, by introducing $m$ new predicates $P_i$ 
and adding the axioms $P_i(x)\leftrightarrow \alpha_i(x)$ and 
$P_j(x,y)\leftrightarrow\alpha_j(x,y)$, if $\alpha_i$ and $\alpha_j$
has one and two free variables respectively, and by defining: 
\begin{align}
 \label{eq:conting-distribution}
q_{\Phi}(\Omega,\bm n) & =\frac{1}{Z}
\sum_{(\bk,\bh)(P_i)=n_i}w(\bk,\bh)\cdot F(\bk,\bh,\{n_{ij{v}}\})
\end{align}
where  $Z={\wfomc(\Omega,w,n)}$ also known as the partition function. 
\end{theorem}
\begin{proof}
The proof is a simple consequence of the fact that all the models agreeing with a  count statistic $\bm N(\Phi,\omega)$ can be 
counted using cardinality constraints which agree with $\bm N(\Phi,\omega)$. Any such cardinality constraint correspond to a specific set of $(\kh)$ vectors. 
Hence, we can express arbitrary probability distributions over count statistics by picking real valued weights for $(\kh)$ vector.
In the following we prove this statement formally:\\
  Since $\Omega$ is a FO$^2$ formula, then we can compute FOMC as
  follows: 
  $$\fomc(\Omega,n) =
  \sum_{\bk,\bh}F(\bk,\bh,\{n_{ijv}\})
  $$
  Let us define $w(\bk,\bh)$ for each $\bk,\bh$ as follows: 
  $$
  w(\bk,\bh)=
  \frac1{F(\bk,\bh,\{n_{ijv}\})}\sum_{\substack{
    \omega\models\Omega \\
    N(\alpha_1,\omega)_1 = (\bk,\bh)(P_1) \\ \dots \\
    N(\alpha_m,\omega)_m = (\bk,\bh)(P_m)}}
p_{\Phi,\Omega}(\omega)
$$
Where $p_{\Phi,\Omega}(\omega)$ is the probability of world $\omega$, under count distribution $q_{\Phi}(\Omega,\bm n)$. Our goal is to 
show that this weight function suffices to express count distributions.
This definition implies that the partition function $Z$ is equal to
1. Indeed:
\begin{align*}
  Z & = \wfomc(\Omega,w,n) \\
    & = \sum_{\bk,\bh}w(\bk,\bh)\cdot F(\bk,\bh,\{n_{ijv}\}) \\
    & = \sum_{\bk,\bh}
        \sum_{\substack{
    \omega\models\Omega \\
    N(\alpha_1,\omega)_1 = (\bk,\bh)(P_1) \\ \dots \\
    N(\alpha_m,\omega)_m = (\bk,\bh)(P_m)}}
  p_{\Phi,\Omega}(\omega) \\
    & = \sum_{\omega\models\Omega}
        \sum_{\substack{
    \bk,\bh \\
    N(\alpha_1,\omega)_1 = (\bk,\bh)(P_1) \\ \dots \\
    N(\alpha_m,\omega)_m = (\bk,\bh)(P_m)}}
  p_{\Phi,\Omega}(\omega) \\
    & = \sum_{\omega\models\Omega}p_{\Phi,\Omega}(\omega) \\ 
& = 1   
\end{align*}
Hence,
\begin{align*}
q_{\Phi}(\Omega,\bm n) & =
\sum_{(\bk,\bh)(P_i)=n_i}
                         F(\bk,\bh,\{n_{ijv}\})\cdot w(\bk,\bh) \\
                       & = \sum_{(\bk,\bh)(P_i)=n_i}
  \sum_{\substack{
    \omega\models\Omega \\
    N(\alpha_1,\omega)_1 = (\bk,\bh)(P_1) \\ \dots \\
    N(\alpha_m,\omega)_m = (\bk,\bh)(P_m)}}
  p_{\Phi,\Omega}(\omega) \\
                       & = 
  \sum_{\substack{
    \omega\models\Omega \\
    N(\alpha_1,\omega)_1 = n_1 \\ \dots \\
    N(\alpha_m,\omega)_m = n_m}}
  p_{\Phi,\Omega}(\omega) \\
\end{align*}
Which is exactly the probability of the worlds agreeing with the count statistic $\bm N(\Phi,\omega)$.
\end{proof}
In \cite{complexMLNkuzelka2020}, authors propose an example for which the probability cannot be expressed using Symmetric-WFOMC 
and obligates the use of complex valued weights. In the following, we present the same example and 
are able to express it's distribution with real valued weights on the $(\kh)$ vector.
\begin{example}
  \label{coins}
In this example we wish to model a sequence of 4 coins being tossed such that 
the probability of getting odd number of heads is zero, and the probability of 
getting even number of heads is uniformly distributed. We introduce a predicate $H(x)$ over a domain of $4$ elements. Notice that such a distribution 
cannot be expressed using symmetric weights, as symmetric weights can only express binomial 
distribution for this language. But we can define weight function on $(\kh)$ vector.  
In this case $\bk=(k_0,k_1)$ such that $k_0+k_1=4$.  Since there are
  no binary predicates we can ignore $\bh$. Intuitively, $k_0$ is the
  number of elements not in $H$ and $k_1$ is the number of elements in
  $H$. If we define the weight function as
$
w(k_0,k_1) = 1 + (-1)^{k_1}
$
by applying \eqref{eq:conting-distribution} we obtain the following
probabilities: 
\begin{align*}
q(\Omega,(4,0))  = \frac{\binom{4}{4}\cdot(1 + 1)}{16} = \frac18 &\hspace*{1cm}  q(\Omega,(3,1))  = \frac{\binom{4}{3}\cdot(1 - 1)}{16}  = 0 \\ 
q(\Omega,(2,2))  = \frac{\binom{4}{2}\cdot(1 + 1)}{16} = \frac34 &\hspace{1cm} q(\Omega,(1,3))  = \frac{\binom{4}{1}\cdot(1 - 1)}{16} = 0 \\ 
q(\Omega,(0,4))  = \frac{\binom{4}{0}\cdot(1 + 1)}{16}  = \frac18 & 
\end{align*}

\end{example}
\begin{arxiv}
We are able to capture count distributions 
without loosing domain liftability or introducing complex or even negative weights,
making the relation between weight functions and probability rather intuitive.
\end{arxiv}


\section{Conclusion}

In this paper we have presented a closed-form formula for FOMC
of universally quantified formulas in FO$^2$ that 
can be computed in polynomial time w.r.t. the size of the domain.
From this, we are able to derive closed-form expression for FOMC 
in FO$^2$ formulas in Scott's Normal Form, extended with cardinality 
constraints. All the formulas are extended to cope
with weighted model counting in a simple way, admitting larger class
of weight functions than symmetric weight functions. 
All the results have been obtained 
without introducing negative or imaginary weights, which makes the
relation between weight functions and probability rather intuitive.



\bibliographystyle{unsrt}  
\bibliography{references}  

\end{document}